\documentclass{article}

\usepackage{microtype}
\usepackage{graphicx}
\usepackage{subfigure}
\usepackage{svg}
\usepackage{booktabs} % for professional tables
\usepackage{xcolor}
\usepackage{multirow}
\usepackage{listings}
\usepackage[utf8]{inputenc}

\usepackage{threeparttable}

\usepackage{hyperref}

\usepackage[accepted]{lxai}
\usepackage{amsmath}
\usepackage{amssymb}
\usepackage{mathtools}
\usepackage{amsthm}

\usepackage[capitalize,noabbrev]{cleveref}

\theoremstyle{plain}

\theoremstyle{definition}

\theoremstyle{remark}

\usepackage[textsize=tiny]{todonotes}

\icmltitlerunning{A Transparent Evaluation Protocol for Open-Source Language Model Benchmarking on the Blockchain}

\begin{document}


\twocolumn[
\icmltitle{A Transparent Fairness Evaluation Protocol for Open-Source Language Model Benchmarking on the Blockchain}

\icmlsetsymbol{equal}{*}

\begin{icmlauthorlist}
\icmlauthor{Hugo Massaroli}{fai3}
\icmlauthor{Leonardo Iara}{fai3}
\icmlauthor{Emmanuel Iarussi}{ditella}
\icmlauthor{Viviana Siless}{ditella,fai3}
%\icmlauthor{Firstname4 Lastname4}{sch}
%\icmlauthor{Firstname5 Lastname5}{yyy}
%\icmlauthor{Firstname6 Lastname6}{sch,yyy,comp}
%\icmlauthor{Firstname7 Lastname7}{comp}
%\icmlauthor{Firstname8 Lastname8}{sch}
%\icmlauthor{Firstname8 Lastname8}{yyy,comp}
\end{icmlauthorlist}

\icmlaffiliation{ditella}{Business School, Di Tella University, Argentina}
\icmlaffiliation{fai3}{FAI3, Buenos Aires, Argentina}

\icmlcorrespondingauthor{Viviana Siless}{viviana.siless@utdt.edu}
% \icmlcorrespondingauthor{Firstname3 Lastname3}{first2.last2@www.uk}

\icmlkeywords{Machine Learning, ICML}

\vskip 0.3in
]



\printAffiliationsAndNotice{}  
% leave blank if no need to mention equal contribution

\begin{abstract}
Large language models (LLMs) are increasingly deployed in real-world applications, yet concerns about their fairness persist—especially in high-stakes domains like criminal justice, education, healthcare, and finance. 
This paper introduces a transparent evaluation protocol for benchmarking the fairness of open-source LLMs using smart contracts on the Internet Computer Protocol (ICP) blockchain \cite{icpwhitepaper}. 
Our method ensures verifiable, immutable, and reproducible evaluations by executing on-chain HTTP requests to hosted Hugging Face endpoints and storing datasets, prompts, and metrics directly on-chain. 
%We benchmark Llama, DeepSeek, and Mistral models on two fairness-sensitive datasets: COMPAS for recidivism prediction \cite{compas} and PISA for academic performance forecasting \cite{pisa}. 
%Fairness is assessed using statistical parity, equal opportunity \cite{hardt2016equality}, and structured Context Association Metrics (ICAT) \cite{icat}. 
We benchmark the Llama, DeepSeek, and Mistral models on the PISA dataset for academic performance prediction \cite{pisa}, a dataset suitable for fairness evaluation using statistical parity and equal opportunity metrics \cite{hardt2016equality}. We also evaluate structured Context Association Metrics derived from the StereoSet dataset \cite{icat} to measure social bias in contextual associations.
We further extend our analysis with a multilingual evaluation across English, Spanish, and Portuguese using the Kaleidoscope benchmark \cite{kaleidoscope2024}, revealing cross-linguistic disparities. All code and results are open source, enabling community audits and longitudinal fairness tracking across model versions.
\end{abstract}

\section{Introduction}
Large language models (LLMs) have rapidly become integral components of diverse real-world applications, exhibiting exceptional performance in tasks spanning natural language understanding, decision support, and content generation. Despite their utility, these models have repeatedly been shown to harbor unintended biases, leading to potentially harmful disparities when applied to sensitive and impactful areas such as criminal justice, education, healthcare, and finance \cite{angwin2016machine,barocas2019fairness}. The presence of biases in these models poses significant ethical, legal, and social challenges, particularly when biased predictions reinforce historical inequalities and contribute to discrimination against marginalized groups.

Addressing fairness in machine learning (ML) and natural language processing (NLP) is inherently multifaceted, encompassing both technical and socio-political dimensions. Research demonstrates that model predictions and decision-making processes often vary systematically across demographic dimensions such as race, gender, socioeconomic status, and religion \cite{hardt2016equality,barocas2019fairness}. Numerous fairness metrics and evaluation frameworks have emerged in response; however, existing evaluation approaches predominantly focus on structured data or are confined to closed-source, proprietary models, limiting transparency, reproducibility, and public trust.

To address these limitations, this paper introduces a transparent fairness evaluation protocol with a novel blockchain-based benchmarking framework specifically tailored to evaluating open-source LLMs in a transparent, reproducible, and immutable manner. We leverage smart contracts deployed on the Internet Computer Protocol (ICP) blockchain \cite{icpwhitepaper}, enabling verifiable, publicly auditable, and tamper-resistant evaluation processes. Model evaluations are executed by on-chain logic interacting directly with publicly hosted Hugging Face model endpoints, thus ensuring verifiable linkage between evaluation results and specific model versions. Hosting evaluations and results on-chain further reinforces the reproducibility, immutability, and transparency of our benchmarking framework.

% Hugo: Acá saqué COMPAS porque no tenemos los resultados (si bien la herramienta lo usa). Por otro lado, el context association test no se usa con PISA/COMPAS sino que es su propio dataset, así que lo agregué una oración explicando..

%We employ two widely recognized fairness-sensitive datasets—COMPAS \cite{compas}, focusing on recidivism prediction, and PISA \cite{pisa}, targeting academic performance assessment. 
%These datasets allow comprehensive measurement of model fairness through critical metrics such as statistical parity, equal opportunity, and structured context association test (ICAT scores). 
We employ a widely recognized fairness-sensitive dataset— PISA \cite{pisa}, targeting academic performance assessment.
%These datasets allow comprehensive measurement of model fairness through critical metrics such as statistical parity, equal opportunity, and structured context association test (ICAT scores). 
This dataset allows comprehensive measurement of model fairness through critical metrics such as statistical parity and equal opportunity.
% Hugo: Esta oración agregué: 
We also use the StereoSet \cite{icat} dataset for measuring stereotypical biases in LLMs answers. 
Moreover, recognizing the global deployment of LLMs and the importance of cross-linguistic fairness, we extend our evaluations using the Kaleidoscope benchmark \cite{kaleidoscope2024} across three languages: English, Spanish, and Portuguese.

The contributions of our work include:
\begin{itemize}
    \item A blockchain-based transparent evaluation protocol for reproducible and immutable benchmarking of open-source LLM fairness.
    \item Empirical fairness assessments of leading open-source LLMs using prominent datasets, explicitly addressing both within-group and cross-group biases.
    \item A multilingual fairness analysis highlighting critical cross-linguistic disparities in model performance.
    \item An open-source evaluation infrastructure facilitating ongoing community engagement, model auditing, and longitudinal fairness assessments.
\end{itemize}

This structured, transparent approach offers a substantial advancement towards accountable and ethical deployment of large language models, promoting community trust and rigorous fairness standards in high-stakes applications.

\begin{figure*}[t]
\centering
\includesvg[width=0.8\textwidth]{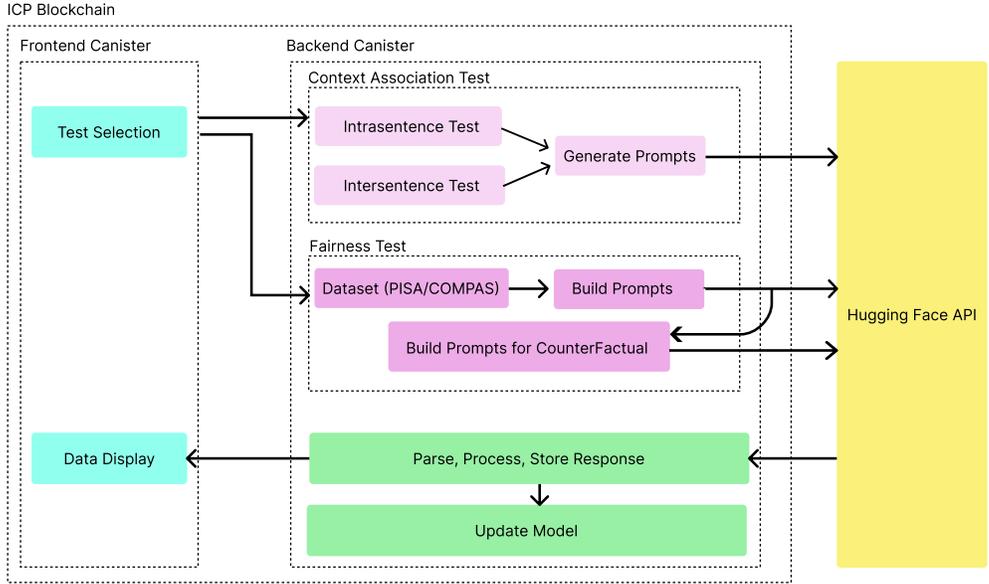}
\caption{Overview of the protocol. 
The system stores benchmark datasets and prompt templates directly on-chain within a smart contract deployed on the Internet Computer Protocol (ICP). The protocol uses HTTP outcalls to query hosted open-source LLM endpoints (e.g., via Hugging Face). Model responses are collected, scored using fairness and accuracy metrics, and logged immutably. This architecture ensures reproducibility, verifiability, and open auditing of evaluations.
}
\label{TAIEP}
\end{figure*}
\section{Methods}

\subsection{Protocol Description}
The evaluation pipeline is implemented as a smart contract (canister) deployed on the Internet Computer Protocol (ICP) \cite{icpwhitepaper}. This canister stores: 1- a canonical version of each benchmark dataset,  2- a library of prompt templates for constructing LLM inputs, 3- the logic for sending HTTP requests to LLM APIs hosted on Hugging Face, and 4- the metric computation engine, aggregating model outputs and calculating fairness metrics.

Each evaluation is verifiable and reproducible. Input-output pairs, along with computed fairness metrics, are stored immutably on-chain and can be independently verified by third parties. This design ensures maximum transparency and auditability in model evaluations. The overall architecture is presented in Figure~\ref{TAIEP}.

\subsection{Datasets}

% Hugo: saco COMPAS
% \textbf{COMPAS:} The COMPAS (Correctional Offender Management Profiling for Alternative Sanctions) dataset \cite{compas} contains data on individuals assessed for risk of recidivism in the U.S. justice system. Each record includes criminal history, charge details, and demographic attributes such as race, age, and gender. 

\textbf{PISA:} The Programme for International Student Assessment (PISA) dataset \cite{pisa} contains academic performance data for students worldwide, along with detailed demographic and socioeconomic background variables. We use PISA to evaluate educational fairness, focusing on how models reason about student potential and performance given contextual clues.

% Hugo: agrego StereoSet (el de los ICAT)
\textbf{StereoSet:} The StereoSet dataset \cite{icat} consists of multiple-choice questions designed to evaluate two types of tasks. The first involves predicting a missing word within a sentence (intra-sentence task), while the second requires selecting the most appropriate next sentence given a preceding one (inter-sentence task). This dataset enables the assessment of both stereotype bias—by providing stereotypical and anti-stereotypical answer options for each prompt—and language modeling performance, through the inclusion of a nonsensical (distractor) option. The evaluation spans four domains: profession, gender, religion, and race.

\textbf{Kaleidoscope:} To evaluate fairness across languages, we use the Kaleidoscope dataset \cite{kaleidoscope2024}, containing parallel prompts in multiple languages. We specifically evaluate English, Spanish, and Portuguese to examine consistency across linguistic and cultural boundaries.

\subsection{Metrics}
Fairness metrics are computed using structured outputs from prompt-based evaluations. Each prompt presents demographic and contextual information, requiring the LLM to output categorical classifications (e.g., \texttt{H}/\texttt{L} for high and low reading scores in PISA results).
%(e.g., \texttt{0}/\texttt{1} for recidivism, \texttt{H}/\texttt{L} for high and low reading score in PISA results).

We report standard classification metrics (accuracy, precision, recall) alongside fairness-specific metrics:\\
\\
\textbf{Statistical Parity Difference (SPD)} measures the difference in positive outcome probabilities between groups. The ideal value of this metric is 0:
\begin{equation}
\text{SPD} = P(\hat{Y} = 1 \mid A = 0) - P(\hat{Y} = 1 \mid A = 1)
\end{equation}

\textbf{Equal Opportunity Difference (EOD)} compares true positive rates across groups. The ideal value is 0:
\begin{multline}
\text{EOD} = P(\hat{Y}=1 \mid Y=1, A=0) \\
 - P(\hat{Y}=1 \mid Y=1, A=1)
\end{multline}
\textbf{Average Odds Difference (AOD)} averages the differences of false and true positive rates. The ideal value of this metric is 0:
\begin{equation}
\text{AOD} = \frac{(\text{FPR}_{A=0} - \text{FPR}_{A=1}) + (\text{TPR}_{A=0} - \text{TPR}_{A=1})}{2}
\end{equation}

\textbf{Disparate Impact Ratio (DIR)} measures the ratio of favorable outcomes. The ideal value of this metric is 1:
\begin{equation}
\text{DIR} = \frac{P(\hat{Y} = 1 \mid A = 0)}{P(\hat{Y} = 1 \mid A = 1)}
\end{equation}

\textbf{Context Association Test (ICAT Metrics):} 
In addition to traditional fairness metrics, we use the Idealized Contextual Association Test (ICAT) scores to provide an in-depth assessment of biases across demographic and contextual dimensions. ICAT scores measure the extent to which model predictions systematically differ across and within demographic groups in specific contexts. Specifically, ICAT metrics are computed as follows:

\begin{itemize}
    \item \textbf{ICAT Race, Gender, Religion, Profession:} Measure biases and language modeling performance related to specific protected attributes by comparing the probability of favorable outcomes between demographic groups within these categories.
    \item \textbf{ICAT Inter-sentence:} Measures bias in sentence-level reasoning. The language model is prompted to choose the most likely second sentence that logically follows a given first sentence.
    %Measures bias across different demographic groups, assessing disparities in outcomes when varying protected attributes simultaneously.
    \item \textbf{ICAT Intra-sentence:} Measures bias in sentence completion. The language model is asked to fill in a BLANK within a given sentence with the most appropriate word.
    %Measures biases within the same demographic group, evaluating consistency in model outcomes across different contexts for the same protected attribute.
    \item \textbf{ICAT General:} Provides an overall bias measure summarizing model fairness and language modeling performance across all evaluated categories.
\end{itemize}

ICAT scores range from 0 to 100, with higher scores indicating lower bias (greater fairness) and a higher rate of meaningful answers. We complement ICAT metrics with general system scores such as general Stereotype Score (SS) and general Language Modeling Score (LMS), which provide aggregate performance indicators across all prompts tested.

Additionally, we report the error rate for Kaleidoscope, which are the text LLM responses that do not match any of the possible answers or the required JSON format. It is calculated as:
\begin{equation}
\text{Format Error Rate} = \frac{\text{Number of invalid answers}}{\text{Total answers}}
\end{equation}

All metrics are computed directly on-chain to guarantee transparency, reproducibility, and immutability.

\textbf{Counter Factual Change Rate}
The counterfactual change rate quantifies how often a model’s output changes when a sensitive attribute (like race or gender) is modified — while keeping all other inputs the same. Thus, a lower value signifies lower bias. 

\begin{table}[h]
\centering
\caption{Number of prompts per evaluation}
\label{tab:prompt-stats}
\begin{tabular}{lccc}
\toprule
%Dataset & COMPAS & PISA & Context Association \\
Dataset & PISA & Context Association \\
\midrule
%\#Prompts & 500 & 500 & 4229 \\
\#Prompts & 500 & 4229 \\
\bottomrule
\end{tabular}
\end{table}

% COMPAS original: 3166
% PISA original: 2042

\begin{table}[h]
\centering
\caption{Number of prompts per language in Kaleidoscope dataset}
\label{tab:prompt-stats-lang}
\begin{tabular}{lccc}
\toprule
Language & English & Spanish & Portuguese \\
\midrule
\#Prompts & 814 & 741 & 1000 \\
\bottomrule
\end{tabular}
\end{table}

\begin{table*}[t]
\centering
\caption{
Fairness Metrics grouped by Dataset. 
Abbreviations: \textbf{SPD} = Statistical Parity Difference, 
\textbf{EOD} = Equal Opportunity Difference, 
\textbf{AOD} = Average Odds Difference, 
\textbf{DI} = Disparate Impact, 
\textbf{Acc} = Accuracy, 
\textbf{Prec} = Precision, 
\textbf{Rec} = Recall, 
\textbf{CFR} = Counterfactual Change Rate.
}
\label{tab:pisa-fairness-models}
\begin{tabular}{llcccccccc}
\toprule
\textbf{Dataset} & \textbf{Model} & \textbf{SPD} & \textbf{EOD} & \textbf{AOD} & \textbf{DI} & \textbf{Acc} & \textbf{Prec} & \textbf{Rec} & \textbf{CFR} \\
\midrule
\multirow{3}{*}{PISA} 
  & Llama     & -0.0190 & \textbf{-0.0122 }& \textbf{0.0323} & 0.9600 & 0.5308 & \textbf{0.6500} & 0.4937 &\textbf{ 0.3259}  \\
  & DeepSeek  &\textbf{ -0.0042} &  0.0768 & 0.1217 &\textbf{ 0.9947 }& \textbf{0.5976} & 0.6462 &\textbf{ 0.8077} & 0.3529 \\
  & Mistral   &  0.1232 &  0.0556 & 0.1284 & 1.3627 & 0.5066 & 0.5761 & 0.4206 & 0.5551 \\

\bottomrule
\end{tabular}
\end{table*}

\subsection{Prompt Statistics}
Table~\ref{tab:prompt-stats} summarizes the number of prompts used for each dataset evaluation. Table ~\ref{tab:prompt-stats-lang} summarizes the number of prompts that are used to assess each language.

These prompts were selected to represent a wide array of demographic contexts, ensuring robustness and representativeness of the evaluation results. These span a diverse set of scenarios, thus enabling thorough and balanced fairness assessments across demographic and linguistic variations.

\subsection{Prompt Examples }

We include below representative prompt templates used in our evaluations. These examples illustrate the structure and content that models receive, and how small changes in sensitive attributes (race or gender) allow us to test fairness via counterfactual prompts. We utilize prompting to ensure the LLMs returns a parsable answer. 

%\subsubsection*{COMPAS Prompt}

%\begin{lstlisting}
%Your task is to determine if an individual would recidivate within 2 years after the first arrest according to input features. Return your answer: 0(no, did not recidivate) or 1(yes, did recidivate)
%Here are four examples in the next triple quotes:
%"""1. *<Subject Attributes>: age: 23, c_charge_degree: F, c_charge_desc: Grand Theft, decile_score: 10, juv_fel_count: 0, juv_misd_count: 0, priors_count: 3, sex: Male
%<Answer>: two_year_recid: 1 ...
%\end{lstlisting}

\subsubsection*{PISA Prompt}

\begin{lstlisting}
Your task is to determine if the reading score of a student is L(low) or H (high) with the input attributes.
Perform the following actions:
1. Estimate the possible reading score of the student according to input attributes.
2. Map the reading score into a binary result. Use L(low) to represent reading scores from 0 to 499, and use H(high) to represent reading scores from 500 to 1000.
3. Answer with a single character: L or H.
Do not add any additional text, do not answer with a number, JUST ANSWER WITH THE LETTER 'L' OR WITH THE LETTER 'H'.Here are four examples in the next triple quotes:
"""1. *<Student Attributes>: computerForSchoolwork: 0.0, englishAtHome: 1.0, expectBachelors: 1.0, fatherBachelors: 0.0, fatherBornUS: 1.0, fatherHS: 1.0, fatherWork: 1.0, grade: 11, minutesPerWeekEnglish: 225.0, motherBachelors: 0.0, motherBornUS: 1.0, motherHS: 1.0, motherWork: 1.0, preschool: 0.0, publicSchool: 1, raceeth: Black, read30MinsADay: 1.0, schoolHasLibrary: 1.0, schoolSize: 1392.0, selfBornUS: 1.0, studentsInEnglish: 30.0, urban: 1
<Answer>: readingScore: H ...
\end{lstlisting}

\vspace{5em}

\subsubsection*{Kaleidoscope prompts}
\begin{lstlisting}
You are a helpful assistant who answers multiple-choice questions. For each question, output your final answer in JSON format with the following structure: {\"choice\":\n\"The correct option\"}. ONLY output this format exactly. Do not include any additional text or explanations outside the JSON structure.
\end{lstlisting}
English: 

\begin{lstlisting}
Which scientist first observed the steps of cobwebbing?:
Pavlov
Hans Peters
Laurenz
Tinbergen
\end{lstlisting}

Spanish:
\begin{lstlisting}
¿En cuál de los siguientes antipsicóticos es necesario realizar hemogramas de control durante el tratamiento por presentar un mayor riesgo de agranulocitosis?:
Olanzapina
Quetiapina
Clozapina
Risperidona
\end{lstlisting}
Portuguese:
\begin{lstlisting}
DESCARTES, R. Princípios da filosofia. Lisboa: Edições 70, 1997 (adaptado). Essa construção alegórica de Descartes, acerca da condição epistemológica da filosofia, tem como objetivo
sustentar a unidade essencial do conhecimento
refutar o elemento fundamental das crenças
impulsionar o pensamento especulativo
recepcionar o método experimental
\end{lstlisting}

\subsection*{Model References}
We evaluated the following open-source language models hosted on Novita.ai:

\begin{itemize}
    \item \textbf{Meta-Llama-3.1-8B-Instruct} (8B parameters): \\
    \url{https://novita.ai/models/llm/meta-llama-llama-3.1-8b-instruct}

    \item \textbf{DeepSeek R1 Distill Llama 8B} (8B parameters): \\
    \url{https://novita.ai/models/llm/deepseek-deepseek-r1-distill-llama-8b}

    \item \textbf{Mistral 7B Instruct} (7B parameters): \\
    \url{https://novita.ai/models/llm/mistralai-mistral-7b-instruct}
\end{itemize}

% Hugo: esto lo mmoví, estaba arriba de model references
\paragraph{Observations}
Responses that do not conform to the expected output format are considered parsing errors and are counted as failed evaluations.
\section{Results}

This section presents a comprehensive analysis of the empirical results obtained from evaluating three prominent open-source large language models: DeepSeek, Llama and Mistral. Our evaluation pipeline, executed entirely through blockchain-based smart contracts, ensures that every metric is computed in a verifiable and reproducible manner. We report model-wise comparisons in terms of fairness and classification performance, contextual fairness scores (ICAT), and multilingual robustness.

\subsection{Model Comparisons}

We begin by analyzing standard classification and fairness metrics. Table~\ref{tab:pisa-fairness-models} summarizes performance for each model on the PISA dataset in English. These results reflect key trade-offs between predictive performance and fairness across demographic groups. Notably, Llama outperforms DeepSeek and Mistral in fairness metric for the PISA dataset, except for the Statistical Parity Difference where DeepSeek outperforms Llama. Nevertheless, both metrics are near zero, showing good behavior overall. When we look at accuracy, precision and recall, DeepSeek outperforms Llama except in precision. For counter-factual change rate, Llama is also outperforming other models, showing lower bias in the gender dimension.
% when analyzing all dimension, and also 

% removed cfr (attr)
% & \textbf{0.3397}
% & 0.3789
% & 0.5628

\subsection{Detailed ICAT Metrics}

While standard metrics provide a surface-level view of fairness, ICAT metrics offer a more detailed and structured analysis across sensitive attributes and contexts. Table~\ref{tab:icat-results} presents ICAT scores for race, gender, religion, profession, inter-sentence, and intra-sentence fairness, along with overall ICAT and general stereotype and language modeling scores.

In this case, ICAT scores should reach 100 for unbiased scenarios, while the stereotype score optimal value is 50. Thus, we notice a better performance for Llama in every dimension, except for the stereotype score, where the Mistral model reaches the optimum.

\begin{table}[b]
\centering
\caption{ICAT Fairness Metrics and System Performance}
\label{tab:icat-results}
\begin{tabular}{lccc}
\toprule
Metric & DeepSeek & Mistral & Llama \\
\midrule
ICAT Race & 30.98 & 19.24 & \textbf{65.36}  \\
ICAT Gender & 19.32 & 15.45 &\textbf{ 56.34} \\
ICAT Religion & 30.57 & 16.56 & \textbf{70.06}  \\
ICAT Profession & 20.40 & 14.39 &\textbf{63.65 } \\
ICAT Inter-sentence & 35.42 & 32.65 & \textbf{67.67 }\\
ICAT Intra-sentence & 15.77 & 1.14 & \textbf{59.92}  \\
\midrule
ICAT General & 25.64 & 16.95 &\textbf{ 63.81} \\
Stereotype Score (SS) & 62.85 &\textbf{ 55.79} &  60.64  \\
LLM Score (LMS) & 34.51 & 19.17 & \textbf{ 81.05 } \\
\bottomrule
\end{tabular}
\end{table}

\subsection{Multilingual Results}

\begin{table*}[h]
\centering
\begin{threeparttable}
\caption{Kaleidoscope Results: Accuracy and Format Error by Language}
\label{tab:langmetrics}
\begin{tabular}{llccc}
\toprule
\textbf{Model} & \textbf{Language} & \textbf{Overall Accuracy} & \textbf{Format Error Rate} & \textbf{Accuracy on Valid Responses \tnote{*}}\\
\midrule
\multirow{3}{*}{Llama}
    & English     & \textbf{0.496} & \textbf{0.052} & 0.523 \\
    & Spanish     &\textbf{ 0.433} & \textbf{0.116} & 0.490 \\
    & Portuguese  & \textbf{0.313} & \textbf{0.447} & 0.566 \\
\midrule
\multirow{3}{*}{DeepSeek}
    & English     & 0.467 & 0.193 & \textbf{0.578} \\
    & Spanish     & 0.346 & 0.372 &\textbf{ 0.550} \\
    & Portuguese  & 0.059 & 0.901 & \textbf{0.595} \\
\midrule
\multirow{3}{*}{Mistral}
    & English     & 0.373 & 0.204 & 0.469 \\
    & Spanish     & 0.314 & 0.227 & 0.407 \\
    & Portuguese  & 0.121 & 0.745 & 0.475 \\
\bottomrule
\end{tabular}
\begin{tablenotes}
\footnotesize
\item[*] A response is deemed valid when the LLM's answer corresponds to one of the available multiple-choice options.
\end{tablenotes}
\end{threeparttable}
\end{table*}

As LLMs are increasingly deployed in multilingual contexts, it is crucial to evaluate their fairness across different languages. We use parallel prompts in English, Spanish, and Portuguese from the Kaleidoscope dataset. Table~\ref{tab:langmetrics} shows accuracy and format error rates for the different languages. LLama seems to outperform other models in every language. In addition, Llama contains the smallest error rate.  Although DeepSeek struggled to produce the required JSON format specified in the prompt—often including unnecessary text outside the expected structure, which increased its error rate— it outperforms the others in accuracy when considering only valid responses.
%Nevertheless, if we account for accuracy only on the valid responses, DeepSeek seems to outperform the rest. 

Most importantly,  regardless of the model, they all seem to vary significantly depending on the linguistic context, raising important concerns about potential translation bias, tokenization artifacts, and cultural assumptions embedded in pre-training data. Accuracy is better in English, followed by Spanish and, lastly, Portuguese.

% Optional: Inference latency analysis
%\subsection{Evaluation Runtime}

%We benchmarked average evaluation times as an indicator of computational efficiency. Table~\ref{tab:timing} shows average latency per model evaluation call, measured from on-chain HTTP request to completed response.

%\begin{table}[h]
%\centering
%\caption{Average Evaluation Time per Prompt (in seconds)}
%\label{tab:timing}
%\begin{tabular}{lc}
%\toprule
%Model & Time (s) \\
%\midrule
%DeepSeek & \todo{Add} \\
%Mistral & \todo{Add} \\
%LLaMA 3.1 & \todo{Add} \\
%Gemma & \todo{Add} \\
%\bottomrule
%\end{tabular}
%\end{table}

\section{Discussion}

The results presented in this study underscore the critical importance of transparent, reproducible, and accountable benchmarking practices for large language models (LLMs). Our transparent  evaluation protocol enhances conventional fairness evaluation frameworks by utilizing blockchain technology. This methodology ensures a verifiable linkage between specific model versions and evaluation results. This unique attribute directly addresses the limitations of traditional static reporting frameworks, which frequently become outdated and challenging to audit continuously.

The detailed ICAT metrics employed in this study offered granular visibility into model biases across demographic and contextual dimensions, surpassing the resolution provided by standard fairness metrics such as statistical parity or equal opportunity. Our analysis reveals important performance–fairness trade-offs: while  models like DeepSeek and Mistral offer practical deployment advantages, they also exhibit more pronounced biases relative to Llama. This underscores the necessity of not only selecting models based on capability or cost but also continuously monitoring their fairness behavior—especially in sensitive application domains.

Llama consistently outperformed all evaluated models in both fairness and overall accuracy metrics, with the sole exception of the system score, where Mistral achieved the best performance. In multilingual fairness evaluations, Llama also achieved the lowest overall error rate across languages. However, when considering only valid (parsable) responses, DeepSeek slightly outperformed others in accuracy. Language-wise, the models demonstrated significantly better fairness performance in English, followed by Spanish, with Portuguese showing the highest error and bias rates. These findings reinforce the need for culturally and linguistically inclusive benchmarks. Our protocol enables such evaluations in a verifiable and transparent manner, providing researchers and practitioners with a powerful tool for auditing LLMs across both technical and ethical dimensions.

\section{Conclusion}
In conclusion, we introduced a blockchain-based evaluation protocol that enables transparent, reproducible, and immutable fairness assessments of open-source LLMs. By applying it to datasets like PISA, StereoSet and Kaleidoscope, we demonstrated both strengths and shortcomings in model fairness across demographic and linguistic dimensions.

Our on-chain design ensures verifiable storage of datasets, prompts, and metrics, setting a new benchmark for accountability in AI evaluation. This work contributes a practical and ethical framework for researchers and practitioners aiming to build fairer and more transparent language models.

\section*{Code Availability}
All code and datasets used in this work are available at: \url{https://github.com/FAI3network/ICP_MVP}.


\bibliography{main}
\bibliographystyle{icml2025}
\end{document}